\documentclass[11pt]{article}

\usepackage[preprint]{acl}
\usepackage{url}
\usepackage{times}
\usepackage{latexsym}

\usepackage[T1]{fontenc}

\usepackage[utf8]{inputenc}

\usepackage{microtype}
\usepackage{placeins}
\usepackage{inconsolata}
\usepackage{makecell}
\usepackage{graphicx}
\usepackage{booktabs}
\usepackage{makecell}
\usepackage{xcolor}
\usepackage{amsmath}
\newcommand{\gain}[1]{\textcolor{teal}{\scriptsize$^{#1}$}}
\newcommand{\drop}[1]{\textcolor{red}{\scriptsize$^{-#1}$}}
\title{Query-Driven Multimodal Information Extraction from Long Documents}

\author{
  \textbf{Yikai Gao\textsuperscript{1}},
  \textbf{Ding Xia\textsuperscript{2}},
  \textbf{Xi Yang\textsuperscript{1,\textdagger}}
\\
  \textsuperscript{1}School of Artificial Intelligence, Jilin University
\\
  \textsuperscript{2}Graduate School of Information Science and Technology, The University of Tokyo
\\
  \texttt{ykgao25@mails.jlu.edu.cn}
  \quad
  \texttt{\{dingxia1995,earthyangxi\}@gmail.com}
\\
  \textsuperscript{\textdagger}Corresponding author
}

\begin{document}
\maketitle
\begin{abstract}
In domain-specific multimodal long documents, images and text jointly convey complex knowledge that cannot be fully captured by plain text alone. However, existing paradigms like DocVQA primarily focus on generating textual answers or localizing evidence regions, rather than outputting query-specific textual attribute values and corresponding images. To address this gap, we propose query-driven image-text joint extraction from long documents, requiring models to output query-requested textual attribute values and corresponding image bounding boxes. Based on challenges related to both user intent and document content, we designed a two-level taxonomy that operates at the query and instance levels. Further, we construct ITJoint, the first high-quality, manually annotated benchmark for this new task, comprising 2,455 pages of domain-specific documents with numerous non-decorative images, 316 queries, and 910 answer instances. Finally, we evaluate representative standalone Vision-Language Models from different providers and further design Q2IT, a multi-agent collaborative framework consisting of three progressively collaborating agents for evidence collection, page selection, and target-image localization. Using a joint evaluation approach that assesses both text extraction and image localization, our experiments show that standalone VLMs struggle with this task, while Q2IT significantly improves performance on ITJoint, although a substantial gap remains toward perfect results.


\end{abstract}

\section{Introduction}
\label{sec:introduction}

As foundation models have learned from massive amounts of web-scale data, their development is increasingly turning toward specialized documents that preserve knowledge accumulated through long-term human practice~\citep{muennighoff2023scaling, villalobos2024position}. Such domain-specific multimodal long documents contain rich textual descriptions and images whose semantics are often tightly coupled and cannot be fully captured by plain text alone. However, transforming this unstructured multimodal information into structured image-text pairs still largely relies on time-consuming manual annotation. Therefore, query-driven image-text joint extraction provides a practical way to satisfy researchers’ targeted needs and to transform domain-specific long documents into high-quality multimodal datasets~\citep{gadre2023datacomp}.

\begin{figure*}[t]
    \centering
    \includegraphics[width=\textwidth]{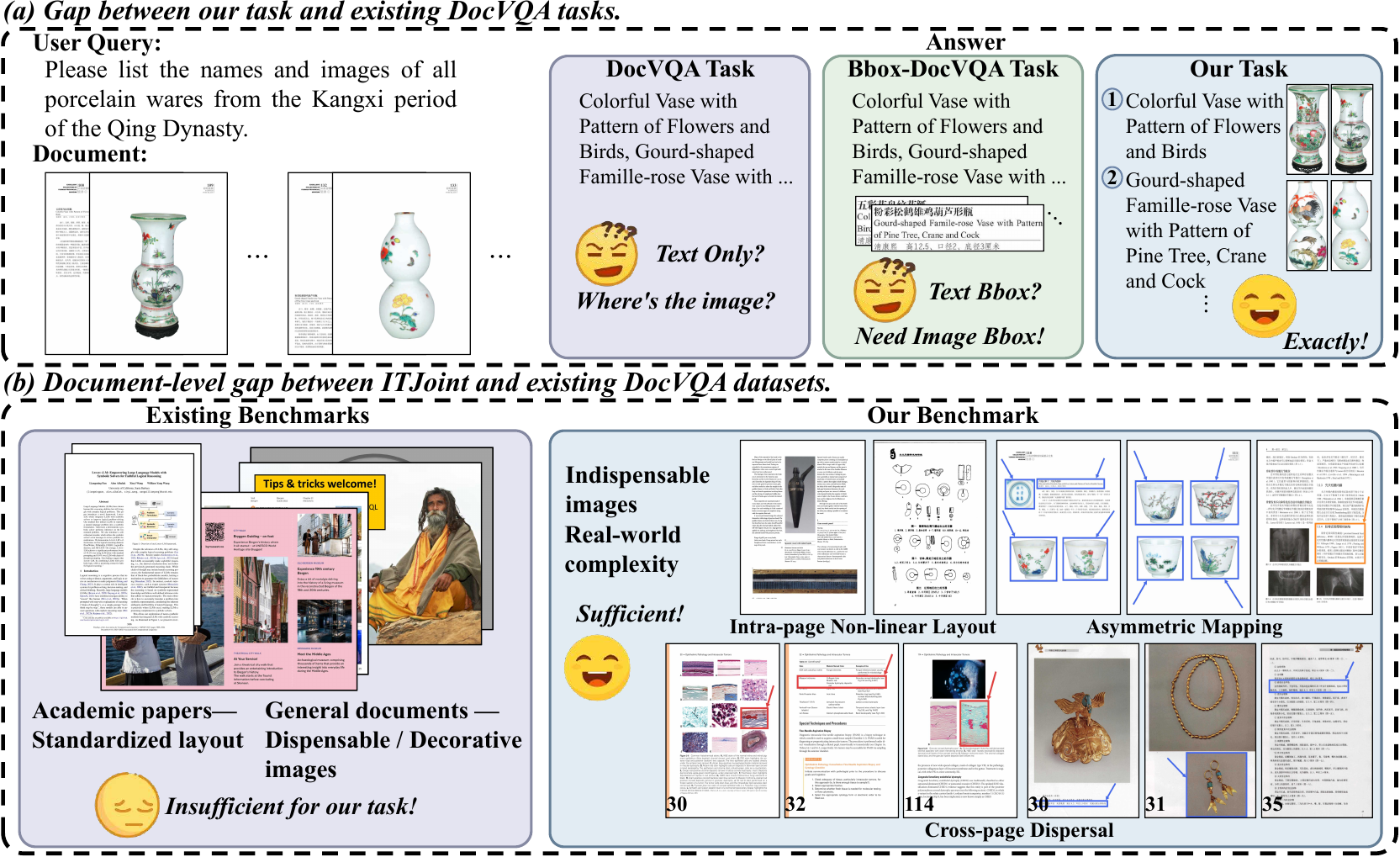}
    \caption{Comparison with existing tasks and datasets. (a) DocVQA tasks mainly return textual answers or evidence regions, while our task requires textual attributes and corresponding images. (b) Existing datasets mainly rely on academic papers or general documents, which are insufficient for evaluating our task.}
    \label{fig:overview}
\end{figure*}


Document visual question answering (DocVQA) has made significant progress~\citep{mathew2021docvqa,tito2023hierarchical,van2023document,ma2024mmlongbench,deng2025longdocurl}, but it remains primarily designed to produce textual answers, even when visual grounding is introduced to localize supporting evidence regions~\citep{yu2025bbox}. Besides, existing image-text extraction methods mostly rely on predefined rules for exhaustive extraction and lack query-driven, on-demand capabilities~\citep{clark2016pdffigures,li2019figure,siegel2018extracting,song2025figex,tang2024pdfchatannotator,pang2024archaeological}. Moreover, existing DocVQA datasets are largely built from academic papers with standardized layouts or general documents where images are mainly decorative or auxiliary, leaving real-world domain-specific long documents with complex layouts and tightly coupled image-text relations underexplored.


To this end, we formally define a new task: query-driven image-text joint extraction from multimodal long documents. Unlike DocVQA, which asks the model to answer a question about a document, our task asks the model to construct query-specific multimodal instances by extracting textual descriptions together with their corresponding target images (Figure~\hyperref[fig:overview]{\ref*{fig:overview}}). 
In other words, given a user query and a long document, the model is required to retrieve the requested textual attributes and return the corresponding image bounding boxes. However, this task poses challenges from both user intent and the document itself. From the perspective of user intent, real-world queries are not limited to clearly specified single-instance queries, but may also involve broadly referential set queries with an uncertain number of targets, requiring methods to balance precise extraction with comprehensive recall. From the document perspective, first, images and text within a page often go beyond linear arrangements and exhibit non-linear and multi-layer nested layouts, making rule-based association methods fragile in real-world domain-specific documents. Second, images and their descriptive text may be dispersed across pages rather than appearing adjacently on the same page, requiring cross-page tracking and aggregation with multi-level context. Third, image-text mappings often go beyond one-to-one correspondences, including one-to-many and text-without-image cases, requiring robust mapping and hallucination suppression. To support systematic study and reliable evaluation, we design a two-level taxonomy at both the query and instance levels, and construct ITJoint, a high-quality manually annotated benchmark for query-driven image-text joint extraction from domain-specific multimodal long documents.

Current approaches to address this task typically involve either standalone Vision-Language Models (VLMs) or carefully designed agent systems tailored for this specific problem. For the VLM-based methods, we selected four state-of-the-art models from different providers. Inspired by existing methods, we also designed Q2IT, a multi-agent collaborative framework that incorporates various modules aimed at enhancing performance on our task. Q2IT consists of three progressively collaborating agents: the Evidence Agent, which retrieves relevant textual information and image clues from the entire document; the Page Agent, which locates the target pages based on these clues; and the Figure Agent, which detects and crops the target images within those pages. 
To evaluate all methods comprehensively, we devised a joint evaluation approach that assesses both image localization and text extraction. Results indicate that current standalone VLMs struggle to handle our task effectively. In contrast, the Q2IT workflow significantly improves performance, demonstrating greater capability in tackling our problem. However, there remains a substantial gap toward achieving perfect results, highlighting considerable room for improvement in both VLMs and multi-agent system designs.


In summary, our contributions are as follows:

\begin{itemize}
    \setlength{\itemsep}{0pt}
    \setlength{\parsep}{0pt}
    \setlength{\parskip}{0pt}
    \item We define query-driven image-text joint extraction from multimodal long documents, requiring models to return query-requested textual attributes and corresponding image bounding boxes. We further analyze its challenges from both user intent and the document itself, and design a two-level taxonomy for systematic evaluation.


    \item We construct ITJoint, the first high-quality manually annotated benchmark for this task. ITJoint targets real-world image-text coupling scenarios in domain-specific long documents, containing 2,455 document pages with an average of 1.77 non-decorative images per page, 316 queries, and 910 answer instances.


    \item We design Q2IT, a task-specific workflow for evidence collection, page selection, and target-image localization, and evaluate it alongside standalone VLMs. Joint assessment shows that Q2IT improves over direct inference, while substantial room remains for further improvement.

\end{itemize}

\section{Related Work}
\paragraph{Document Visual Question Answering and Benchmarks.}
DocVQA aims to answer user questions from document images~\citep{mathew2021docvqa}. Related benchmarks have evolved from single-page to multi-page documents~\citep{tito2023hierarchical}, and further to multidomain, visually rich, and long-context multimodal document settings~\citep{van2023document,ma2024mmlongbench,deng2025longdocurl}. Methods have also progressed from OCR- and layout-based document understanding models~\citep{xu2021layoutlmv2,huang2022layoutlmv3} to OCR-free models~\citep{kim2021donut} and document-oriented multimodal large language models~\citep{ye2023mplug}. BBox-DocVQA further introduces evidence bounding boxes for answer grounding~\citep{yu2025bbox}. Nevertheless, existing DocVQA still centers on textual answers: even with bounding boxes, localization mainly serves as evidence grounding rather than extracting query-targeted textual attributes and corresponding image coordinates. Existing benchmarks are also dominated by structured academic papers or general documents where images are mostly auxiliary. 
Therefore, existing DocVQA methods and benchmarks remain insufficient for query-driven image-text joint extraction.

\paragraph{Image-Text Pair Extraction from Documents.}
Another line of work extracts images and associated text from documents, including figure-caption extraction in scientific papers and multimodal data collection from domain-specific catalogs. Methods such as PDFFigures 2.0~\citep{clark2016pdffigures}, PDFigCapX~\citep{li2019figure}, and DeepFigures~\citep{siegel2018extracting} extract figures, captions, and coordinates from academic papers, while later work studies subfigure-subcaption alignment in compound figures~\citep{song2025figex}. For domain-specific catalogs, PDFChatAnnotator~\citep{tang2024pdfchatannotator} supports multimodal data collection from PDF-format catalogs through human-LLM collaboration, and VLM-based methods have been used to collect artifact images and attributes from archaeological catalogs~\citep{pang2024archaeological}. However, existing methods typically rely on predefined targets, fixed fields, or human-collaborative workflows, and cannot select target instances on demand from open-ended user queries. 

\paragraph{Multi-Agent Frameworks for Complex Tasks.}
Multi-agent frameworks have been widely used to decompose complex tasks into collaborative subprocesses. General systems such as CAMEL~\citep{li2023camel}, AutoGen~\citep{wu2024autogen}, and MetaGPT~\citep{hong2024metagpt} support planning, execution, and refinement through role specialization, conversation, or procedural intermediate artifacts. In downstream applications, P2P~\citep{sun2025p2p} and Paper2Poster~\citep{pang2026paper2poster} decompose paper-to-poster generation into content parsing, visual organization, layout planning, and iterative refinement; AutoPrep~\citep{fan2024autoprep} uses multiple agents for question-driven data preparation; and oracle bone script interpretation adopts an agent-driven multimodal knowledge augmentation framework for domain-specific reasoning~\citep{zhang2026specializing}. Inspired by these studies, we model image-text joint extraction from multimodal long documents as a staged collaborative process with multi-agents. 
\section{Task Definition and Benchmark}
\subsection{Task Formulation}
Given a user query $q$ and a long document $D=\{p_i\}_{i=1}^{N}$, the model is required to return a set of answer instances:
\begin{equation}
\mathcal{R}=\{z_j\}_{j=1}^{J}, \quad z_j=(t_j,v_j).
\label{eq:task_output}
\end{equation}
Each answer instance $z_j$ consists of a textual attribute value $t_j$ satisfying the query condition and an associated image set $v_j$:
\begin{equation}
v_j=\{(\pi_{j,m}, b_{j,m})\}_{m=1}^{M_j},
\label{eq:image_boxes}
\end{equation}
where $\pi_{j,m}$ and $b_{j,m}$ denote the page index and bounding box of the $m$-th associated image, respectively. When textual evidence for an instance exists but no corresponding image is present, the system must still return $t_j$ and explicitly output an empty image set, i.e., $v_j=\emptyset$, rather than omitting the instance or hallucinating an image match.

\begin{figure}[t]
    \centering
    \includegraphics[width=\columnwidth]{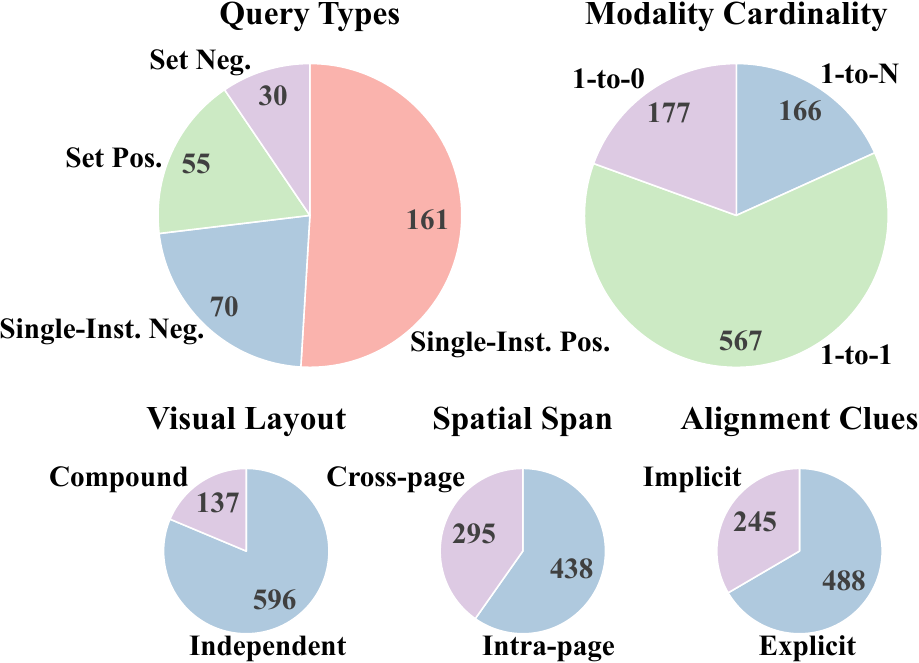}
    \caption{Distributions of ITJoint under the query-level and instance-level taxonomies.}
    \label{fig:dataset_distribution}
\end{figure}

\subsection{Challenge Taxonomy}

To systematically characterize the challenges of producing $\mathcal{R}$ and support fine-grained evaluation, we define a two-level taxonomy over $q$ and $z_j=(t_j,v_j)$.

\paragraph{Query-level taxonomy.}

To characterize challenges from user intent, we define two query-level categories. \textit{Single-instance queries} require $J=1$ and request image-text information for one specific answer instance. \textit{Set queries} require all answer instances satisfying a given condition, where $J$ is not known in advance. Based on this distinction, we further refine the conventional negative-query setting in existing benchmarks~\citep{ma2024mmlongbench,deng2025longdocurl}, where unanswerability is caused by the complete absence of textual evidence, into a task-specific negative setting where textual evidence exists but the corresponding image is missing.

Specifically, positive queries require their answer instances to contain both the requested textual attributes and corresponding images. Negative queries differ across the two query types. For single-instance queries, the target instance has textual evidence but no corresponding image, resulting in a negative answer instance with $v_j=\emptyset$. For set queries, image-bearing and text-only instances coexist in $\mathcal{R}$, requiring the system to extract valid image-text pairs while explicitly identifying instances with $v_j=\emptyset$.

\paragraph{Instance-level taxonomy.}

To characterize challenges from the document itself, we annotate each $z_j=(t_j,v_j)$ with fine-grained labels. For \textit{modality cardinality}, instances are labeled according to $M_j$ in Eq.~\ref{eq:image_boxes}: 1-to-0 if $M_j=0$, 1-to-1 if $M_j=1$, and 1-to-N if $M_j>1$. Instances labeled as 1-to-0 contain textual evidence but no corresponding image and are therefore negative answer instances, while 1-to-1 and 1-to-N instances are positive answer instances.

For positive answer instances, we further annotate three dimensions that describe how $t_j$ and $v_j$ are associated in the document. For \textit{visual layout}, instances are labeled as independent or compound, distinguishing whether the target image in $v_j$ is a standalone image or is nested with other images in a compound image. For \textit{spatial span}, instances are labeled as intra-page or cross-page according to whether the textual evidence and all image pages in $v_j$ together involve a single page or multiple pages. For \textit{alignment clues}, instances are labeled as explicit or implicit. Explicit alignment indicates that the document contains clear textual markers that directly establish the correspondence between $t_j$ and $v_j$, whereas implicit alignment indicates that such markers are missing or insufficient, requiring the system to infer the correspondence using additional signals such as layout position, reading order, and visual content. Appendix~\ref{app:taxonomy_examples} provides visual examples of these instance-level labels.

\begin{table}[t]
\centering
\caption{Statistics of our collected ITJoint.}
\label{tab:dataset_statistics}
\small
\begin{tabular}{@{}p{0.68\columnwidth}r@{}}
\toprule
\textbf{Statistic} & \textbf{Number} \\
\midrule
\textbf{Documents} & \\
\quad Total pages & 2,455 \\
\quad Avg. pages & 59.88 \\
\quad Max / Min pages & 178 / 22 \\
\quad Avg. images / page & 1.77 \\
\midrule
\textbf{Set Queries} & \\
\quad Total instances & 679 \\
\quad Avg. instances / query & 7.99 \\
\quad Max / Min instances / query & 62 / 1 \\
\quad Pos. / neg. instances in neg. sets & 148 / 107 \\
\midrule
\textbf{1-to-N Instances} & \\
\quad Total images & 450 \\
\quad Avg. images / instance & 2.71 \\
\quad Max / Min images / instance & 13 / 2 \\
\bottomrule
\end{tabular}
\end{table}

\subsection{ITJoint Benchmark}

\paragraph{Document Collection.}
We collect long documents from multiple domain-specific fields, including medicine, agriculture, archaeology, biology, and art. These documents contain diverse image-text layouts, and include rich non-decorative core images. In addition, the selected documents cover both English and Chinese, enabling evaluation of models' applicability across different language settings. Document examples are provided in Appendix~\ref{app:document_examples}.


\begin{figure*}[t]
    \centering
    \includegraphics[width=\textwidth]{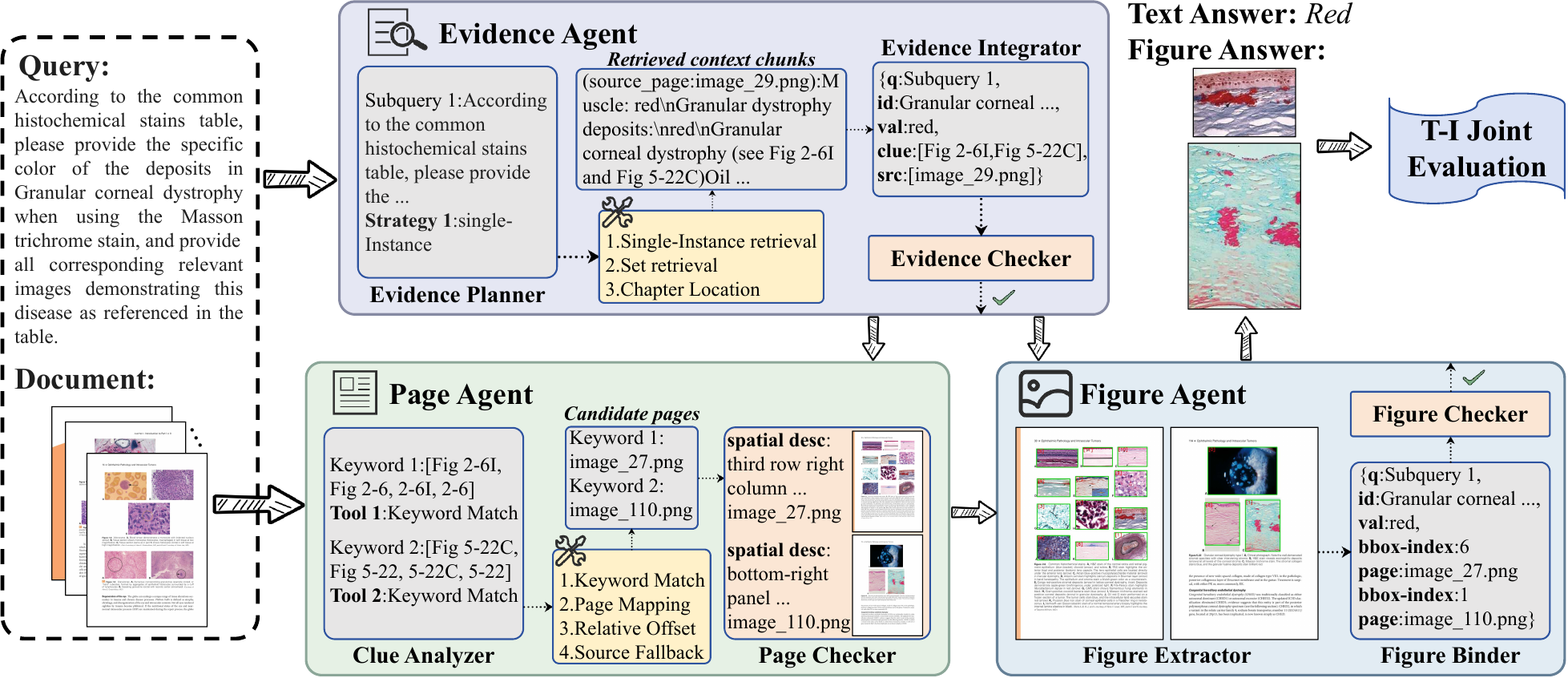}
    \caption{Overview of Q2IT: a three-agent framework for query-driven image-text joint extraction. The Evidence Agent performs evidence collection by collecting textual attribute values and image-reference clues, the Page Agent performs page localization by identifying target pages and coarse spatial descriptions, and the Figure Agent completes image-text binding by selecting target image regions and producing final answer instances.}
    \label{fig:method}
\end{figure*}

\paragraph{Annotation Workflow.}
We adopt a human-LLM collaborative annotation workflow. First, we extract the plain text from long documents and use a large language model to generate candidate queries in batches according to the query-level taxonomy. To improve annotation efficiency, we develop \textbf{ITLabel}, a visual annotation tool specifically designed for this task. During formal annotation, annotators use ITLabel to manually clean candidate queries and annotate answers following predefined annotation guidelines.

Specifically, annotators read through the document page by page to identify all answer instances satisfying the query conditions, with particular care taken to avoid missing instances for set queries whose target count is uncertain. They also assign the corresponding query-level and instance-level labels to each answer instance. During this process, candidate queries that do not satisfy the predefined rules are either filtered out or revised according to the actual document content. More details about the ITLabel annotation tool, annotation guidelines, and annotation examples are provided in Appendix~\ref{app:annotation}.




\paragraph{Dataset Statistics.}
Figure~\ref{fig:dataset_distribution} and Table~\ref{tab:dataset_statistics} summarize the statistics of ITJoint. ITJoint contains 2,455 domain-specific document pages, with an average document length of 59.88 pages and 1.77 non-decorative images per page on average. It includes 316 queries and 910 answer instances. Among them, set queries correspond to 679 answer instances, with an average of 7.99 instances per set query and a maximum of 62 instances for a single query. For 1-to-N cases, the 166 instances involve 450 images in total, with an average of 2.71 images per instance and a maximum of 13 images.

\section{Method}

As shown in Figure~\ref{fig:method}, given a query $q$ and a document $D$, Q2IT progressively constructs the answer set $\mathcal{R}$ through three stages: evidence collection, page selection, and target-image localization, handled by the Evidence Agent, Page Agent, and Figure Agent, respectively.

Before the agent workflow, Q2IT renders each page of $D$ as a 144-DPI image, extracts page-level text with an LLM, and segments the text into page-indexed chunks. These chunks are indexed by sparse and dense retrieval and clustered with a Gaussian Mixture Model (GMM) to support set-query recall. More implementation details are provided in Appendix~\ref{app:implementation_details}.

\subsection{Evidence Agent}

The Evidence Agent extracts candidate textual values and image-reference clues. The \textbf{Evidence Planner} first decomposes $q$ into subqueries $\{\tilde{q}_a\}_{a=1}^{A}$ and selects retrieval strategies according to query intent. For single-instance queries, it retrieves the top-$k$ relevant chunks through hybrid retrieval. For set queries, it retrieves the top-$k$ chunks, selects seed chunks with an LLM, and expands them to the corresponding GMM clusters. If a subquery specifies a section range, the system first locates section boundaries; if this fails, it falls back to retrieval.

The \textbf{Evidence Integrator} converts retrieved contexts into a subquery-conditioned evidence collection:
\begin{equation}
\begin{gathered}
\mathcal{E}=\{(\tilde{q}_a,\mathcal{E}_a)\}_{a=1}^{A}, \quad
\mathcal{E}_a=\{e_{a,r}\}_{r=1}^{n_a}, \\
e_{a,r}=(\mathrm{id}_{a,r}, \mathrm{val}_{a,r}, 
\mathrm{clue}_{a,r}, \mathrm{src}_{a,r}).
\end{gathered}
\label{eq:evidence}
\end{equation}
Here, $\mathrm{id}_{a,r}$ denotes the candidate instance identifier, $\mathrm{val}_{a,r}$ is the candidate textual value, $\mathrm{clue}_{a,r}$ contains image-reference clues, and $\mathrm{src}_{a,r}$ records source pages. The \textbf{Evidence Checker} removes irrelevant evidence and triggers supplementary retrieval when textual values are missing or incomplete.

\subsection{Page Agent}
\label{sec:page_agent}

The Page Agent locates candidate target pages for each evidence item. It follows a clue-driven strategy: explicit labels are localized through keyword matching, printed page numbers through page-index mapping, relative references through offset computation, and intra-page spatial descriptions by directly using the evidence page. If clue-based localization fails, the system falls back to entity-based full-document matching and then to the source pages in $\mathrm{src}_{a,r}$.

The \textbf{Page Checker} visually verifies candidate pages and their neighboring pages, producing verified target pages and coarse spatial descriptions:
\begin{equation}
\begin{gathered}
\mathcal{L}=\{\mathcal{L}_a\}_{a=1}^{A}, \quad
\mathcal{L}_a=\{\ell_{a,r}\}_{r=1}^{n_a}, \\
\ell_{a,r}=\{(\rho_{a,r,u}, \mathrm{loc}_{a,r,u})\}_{u=1}^{U_{a,r}}.
\end{gathered}
\label{eq:page_output}
\end{equation}
Here, $\rho_{a,r,u}$ denotes a verified target page, and $\mathrm{loc}_{a,r,u}$ denotes the coarse spatial description on that page for subsequent image selection.

\subsection{Figure Agent}

The Figure Agent adopts a detect-then-select strategy to extract target images and complete image-text binding. For each verified page $\rho_{a,r,u}$, the \textbf{Figure Extractor} uses Grounding DINO~\citep{liu2024grounding} to obtain candidate boxes, followed by non-maximum suppression, area filtering, and validity filtering.

The \textbf{Figure Binder} reformulates image binding as in-page candidate selection. Candidate boxes are numbered on the original page, and the model selects the target box using $e_{a,r}$, $\tilde{q}_a$, and $\mathrm{loc}_{a,r,u}$. This preserves page-level context such as captions, labels, and layout relations.

Finally, the \textbf{Figure Checker} removes duplicate instances and merges complementary results from different subqueries. It combines the verified textual value and selected image boxes into final answer instances:
\begin{equation}
\mathcal{R}=\{z_j\}_{j=1}^{J}, \quad z_j=(t_j,v_j).
\label{eq:final_output}
\end{equation}

\section{Experiments}

\subsection{Experimental Settings}

\paragraph{Metrics.}
Since this task requires both image-region localization and textual attribute extraction, we evaluate models from three perspectives: image localization, text extraction, and image-text joint correctness. For each query, we first greedily match predicted instances with gold instances that contain images based on instance-level bounding-box overlap. After all image-bearing instances are matched, the remaining 1-to-0 gold instances are matched with the remaining predictions based on textual equivalence. For a matched 1-to-0 instance, image localization is considered correct only if the prediction also outputs an empty image set; otherwise, it is counted as an image-text alignment hallucination.

For image localization, we use an IoU threshold of $\tau=0.75$ and define two settings: Strict and Soft. Strict requires the number of predicted image boxes to exactly match the number of gold boxes, with every matched box satisfying the IoU threshold. Soft relaxes the cardinality constraint, allowing a predicted instance to contain fewer boxes than the gold instance, but requiring every predicted box to match a gold box of the same instance with IoU no lower than $\tau$. For text extraction, we adopt the LLM-as-a-Judge paradigm and assign each instance a binary correctness label. The joint metric requires both image localization and text extraction to be correct. We use Strict Joint as the main evaluation setting, reporting macro-F1 at the query level and accuracy at the instance level. Soft results are provided in Appendix~\ref{app:soft_results}.

\paragraph{Baselines.}
We select four advanced multimodal large language models as baselines. For closed-source models, we use GPT-5.4~\citep{openai2026gpt54} and Gemini-3.1-Pro-Preview~\citep{google2026gemini31propreview}. For open-source models, we use GLM-4.5V~\citep{hong2025glm} and Qwen3-VL-235B-A22B-Instruct~\citep{bai2025qwen3}. In the experiments, we refer to them as GPT, Gemini, GLM, and Qwen, respectively. Each baseline receives all page images of the document together with the query in a single call, and directly outputs structured answer instances. Each document page is rendered as a 144-DPI PNG image. When the number of pages exceeds the model's single-call image limit, we follow the stitching strategy~\citep{ma2024mmlongbench}, arranging pages into composite images with a three-column grid and mapping the output coordinates back to the original page coordinates. Due to API limitations, GPT, GLM, and Qwen support up to 50 images per call, while Gemini supports up to 16. We set the temperature of all models to 0.0 for reproducibility.

\begin{table*}[t]
\centering
\caption{Main results on ITJoint under the Strict Joint setting. We evaluate VLMs in two settings: standalone direct inference and as the backbone of Q2IT.}
\label{tab:main_results}
\scriptsize
\setlength{\tabcolsep}{3.6pt}
\renewcommand{\arraystretch}{1.12}
\resizebox{\textwidth}{!}{
\begin{tabular}{lccccccccccccc}
\toprule
\textbf{Model}
& \makecell{\textbf{Single}\\\textbf{Pos.}}
& \makecell{\textbf{Single}\\\textbf{Neg.}}
& \makecell{\textbf{Set}\\\textbf{Pos.}}
& \makecell{\textbf{Set}\\\textbf{Neg.}}
& \textbf{1-to-0}
& \textbf{1-to-1}
& \textbf{1-to-N}
& \textbf{Indep.}
& \textbf{Comp.}
& \textbf{Intra}
& \textbf{Cross}
& \textbf{Exp.}
& \textbf{Imp.} \\
\midrule
\multicolumn{14}{c}{\textbf{Standalone VLMs}} \\
\midrule
Gemini-3.1-Pro-Preview
& \textbf{0.447} & \textbf{0.714} & \textbf{0.400} & \textbf{0.545}
& \textbf{0.744} & \textbf{0.482} & \textbf{0.291}
& \textbf{0.498} & \textbf{0.182}
& \textbf{0.502} & \textbf{0.344}
& \textbf{0.452} & \textbf{0.412} \\

GPT-5.4
& 0.099 & 0.393 & 0.104 & 0.310
& 0.541 & 0.113 & 0.024
& 0.113 & 0.007
& 0.084 & 0.105
& 0.074 & 0.131 \\

Qwen3-VL-235B-A22B-Instruct
& 0.249 & 0.309 & 0.225 & 0.298
& 0.387 & 0.230 & 0.100
& 0.226 & 0.093
& 0.235 & 0.149
& 0.231 & 0.147 \\

GLM-4.5V
& 0.074 & 0.304 & 0.046 & 0.158
& 0.273 & 0.034 & 0.078
& 0.050 & 0.015
& 0.046 & 0.041
& 0.043 & 0.045 \\

\midrule
\multicolumn{14}{c}{\textbf{VLMs within Q2IT}} \\
\midrule
Gemini-3.1-Pro-Preview
& \textbf{0.692}\gain{+0.244}
& \textbf{0.943}\gain{+0.229}
& 0.535\gain{+0.136}
& \textbf{0.714}\gain{+0.169}
& \textbf{0.876}\gain{+0.132}
& \textbf{0.563}\gain{+0.081}
& \textbf{0.476}\gain{+0.185}
& \textbf{0.577}\gain{+0.080}
& \textbf{0.394}\gain{+0.212}
& \textbf{0.539}\gain{+0.036}
& \textbf{0.549}\gain{+0.206}
& 0.572\gain{+0.120}
& \textbf{0.486}\gain{+0.074} \\

GPT-5.4
& 0.608\gain{+0.508}
& 0.805\gain{+0.412}
& \textbf{0.571}\gain{+0.466}
& 0.691\gain{+0.381}
& 0.802\gain{+0.262}
& 0.534\gain{+0.422}
& 0.452\gain{+0.428}
& 0.545\gain{+0.433}
& 0.387\gain{+0.380}
& 0.530\gain{+0.445}
& 0.495\gain{+0.390}
& \textbf{0.574}\gain{+0.500}
& 0.400\gain{+0.269} \\

Qwen3-VL-235B-A22B-Instruct
& 0.476\gain{+0.226}
& 0.623\gain{+0.314}
& 0.422\gain{+0.197}
& 0.407\gain{+0.109}
& 0.528\gain{+0.141}
& 0.382\gain{+0.152}
& 0.321\gain{+0.221}
& 0.395\gain{+0.168}
& 0.250\gain{+0.157}
& 0.388\gain{+0.153}
& 0.338\gain{+0.189}
& 0.410\gain{+0.178}
& 0.286\gain{+0.139} \\

GLM-4.5V
& 0.485\gain{+0.410}
& 0.752\gain{+0.448}
& 0.403\gain{+0.358}
& 0.557\gain{+0.398}
& 0.644\gain{+0.371}
& 0.388\gain{+0.355}
& 0.277\gain{+0.199}
& 0.383\gain{+0.332}
& 0.277\gain{+0.263}
& 0.395\gain{+0.349}
& 0.315\gain{+0.275}
& 0.416\gain{+0.373}
& 0.257\gain{+0.212} \\
\bottomrule
\end{tabular}
}
\end{table*}

\begin{table}[t]
\centering
\caption{Decomposed results on Text, Page, Image, and Joint metrics.}
\label{tab:decomposed_results}
\scriptsize
\setlength{\tabcolsep}{4pt}
\renewcommand{\arraystretch}{1.12}
\resizebox{\columnwidth}{!}{
\begin{tabular}{lcccc}
\toprule
\textbf{Model} & \textbf{Text} & \textbf{Page} & \textbf{Image} & \textbf{Joint} \\
\midrule
\multicolumn{5}{c}{\textbf{Standalone VLMs}} \\
\midrule
Gemini-3.1-Pro-Preview & \textbf{0.615} & \textbf{0.881} & \textbf{0.518} & \textbf{0.507} \\
GPT-5.4 & 0.306 & 0.808 & 0.187 & 0.185 \\
Qwen3-VL-235B-A22B-Instruct & 0.414 & 0.776 & 0.268 & 0.263 \\
GLM-4.5V & 0.243 & 0.629 & 0.131 & 0.128 \\
\midrule
\multicolumn{5}{c}{\textbf{VLMs within Q2IT}} \\
\midrule
Gemini-3.1-Pro-Preview & \textbf{0.787}\gain{+0.172}
       & \textbf{0.908}\gain{+0.027}
       & \textbf{0.793}\gain{+0.275}
       & \textbf{0.722}\gain{+0.215} \\
GPT-5.4 & 0.763\gain{+0.457}
       & 0.872\gain{+0.063}
       & 0.725\gain{+0.538}
       & 0.653\gain{+0.468} \\
Qwen3-VL-235B-A22B-Instruct & 0.669\gain{+0.255}
       & 0.786\gain{+0.011}
       & 0.553\gain{+0.285}
       & 0.492\gain{+0.229} \\
GLM-4.5V & 0.668\gain{+0.425}
       & 0.756\gain{+0.126}
       & 0.595\gain{+0.464}
       & 0.537\gain{+0.409} \\
\bottomrule
\end{tabular}
}
\end{table}

\begin{table}[t]
\centering
\caption{Results on 1-to-0 instances.}
\label{tab:negative_results}
\scriptsize
\setlength{\tabcolsep}{4pt}
\renewcommand{\arraystretch}{1.12}
\resizebox{0.78\columnwidth}{!}{
\begin{tabular}{lccc}
\toprule
\textbf{Model} & \textbf{Text Acc.} & \textbf{Joint Acc.} & \textbf{Gap}$\downarrow$ \\
\midrule
\multicolumn{4}{c}{\textbf{Standalone VLMs}} \\
\midrule
Gemini-3.1-Pro-Preview & \textbf{0.849} & \textbf{0.744} & \textbf{0.105} \\
GPT-5.4 & 0.767 & 0.541 & 0.226 \\
Qwen3-VL-235B-A22B-Instruct & 0.714 & 0.387 & 0.327 \\
GLM-4.5V & 0.597 & 0.273 & 0.324 \\
\midrule
\multicolumn{4}{c}{\textbf{VLMs within Q2IT}} \\
\midrule
Gemini-3.1-Pro-Preview & \textbf{0.904}\gain{+0.055}
       & \textbf{0.876}\gain{+0.132}
       & \textbf{0.028} \\
GPT-5.4 & 0.887\gain{+0.120}
       & 0.802\gain{+0.262}
       & 0.085 \\
Qwen3-VL-235B-A22B-Instruct & 0.778\gain{+0.064}
       & 0.528\gain{+0.141}
       & 0.250 \\
GLM-4.5V & 0.757\gain{+0.160}
       & 0.644\gain{+0.371}
       & 0.113 \\
\bottomrule
\end{tabular}
}
\end{table}

\subsection{Direct and Workflow-Based Results}
\label{sec:task_difficulty}

Table~\ref{tab:main_results} compares two representative ways to address the proposed task: direct inference with standalone VLMs and workflow-based extraction with Q2IT.

\paragraph{Standalone VLMs.}
Under direct inference, different VLMs show clear performance gaps. Gemini performs best across all query-level categories and most instance-level categories, indicating the strongest direct capability for long-document image-text joint extraction. Specifically, Gemini achieves 0.447 on single positive queries, 0.400 on set positive queries, 0.482 on 1-to-1 instances, and 0.291 on 1-to-N instances, while reaching 0.744 on 1-to-0 instances.

Qwen is the strongest open-source standalone model, especially on positive image-bearing categories. It achieves 0.230 on 1-to-1 instances and 0.100 on 1-to-N instances, outperforming GPT and GLM in these settings. GPT shows a more imbalanced pattern: it performs relatively better on negative or 1-to-0 cases, but struggles with positive instances requiring image localization, achieving only 0.113 on 1-to-1 and 0.024 on 1-to-N instances. GLM is the weakest overall, remaining below 0.10 on most positive image-bearing categories. Overall, Gemini leads by a clear margin among standalone VLMs, Qwen is more stable among open-source models, while GPT and GLM show substantial limitations in directly producing complete image-text answer instances.

\paragraph{Q2IT workflow.}
Under the Q2IT setting, model differences remain clear, but their capability profiles change. Gemini is the most stable backbone, achieving 0.692 on single positive queries, 0.943 on single negative queries, 0.876 on 1-to-0 instances, 0.563 on 1-to-1 instances, and 0.476 on 1-to-N instances. It also obtains the best or near-best results on difficult categories such as compound, cross-page, and implicit-alignment instances, showing strong overall reliability as the workflow backbone.

GPT becomes highly competitive within Q2IT. It achieves the best score on set positive queries with 0.571, and reaches 0.534 on 1-to-1 instances and 0.452 on 1-to-N instances, close to Gemini. It also performs strongly on explicit-alignment instances, reaching 0.574.

Open-source models still lag behind Gemini and GPT under Q2IT, but they show different strengths. Qwen performs slightly better than GLM on set positive queries, 1-to-N instances, and implicit alignment, suggesting an advantage in multi-instance and multi-image scenarios. GLM performs better on negative, 1-to-0, and some explicit-clue categories, indicating relatively stronger text-only or no-image judgment. Overall, closed-source models remain stronger in the Q2IT setting, with Gemini being more stable and GPT standing out on set queries and explicit alignment, while open-source models exhibit distinct but still limited capabilities.

Despite these gains, most categories remain below 0.60 even under Q2IT, indicating substantial room for improving both model capability and workflow design.
\subsection{Decomposed Analysis}
\label{sec:sources_of_difficulty}

Table~\ref{tab:decomposed_results} breaks down performance into query-level macro-F1 scores for Text, Page, Image, and Joint metrics, while Table~\ref{tab:negative_results} reports Text Acc. and Joint Acc. on 1-to-0 instances.

\paragraph{From Page Localization to Image Localization.}
Page scores are much higher than Image scores, especially for standalone VLMs. Gemini drops from 0.881 on Page to 0.518 on Image, while GPT drops from 0.808 to 0.187. This suggests that models can often reach the correct pages but fail to localize the target images. Q2IT substantially improves Image scores, from 0.518 to 0.793 for Gemini and from 0.187 to 0.725 for GPT, while the gains on Page are relatively smaller. This indicates that a key benefit of the workflow lies in converting page-level evidence into accurate bbox-level image localization.

\paragraph{From Single-Side Correctness to Joint Correctness.}
Q2IT improves both Text and Image, but Joint remains lower than either metric. For example, Gemini with Q2IT achieves 0.787 on Text and 0.793 on Image, but 0.722 on Joint; GPT achieves 0.763, 0.725, and 0.653, respectively. This gap shows that correct text extraction and correct image localization do not necessarily form a correct answer instance. The remaining difficulty lies in binding the textual attribute to the corresponding image, especially in multi-image, compound-layout, and implicit-alignment cases.

\paragraph{Text-only Cases.}
The 1-to-0 setting further exposes hallucinated image binding. These instances contain textual evidence but no corresponding image, so the gap between Text Acc. and Joint Acc. reflects cases where the model identifies the textual instance but incorrectly assigns an image. Q2IT reduces this gap, such as from 0.226 to 0.085 for GPT and from 0.324 to 0.113 for GLM, but does not eliminate it. This shows that the task also requires models to recognize when no image-text association should be made. A qualitative example is provided in Appendix~\ref{app:hallucination_example}.

\subsection{Additional Ablation Analysis}
\label{sec:additional_ablation}

To further investigate the challenges discussed above, we provide additional ablation studies in Appendix~\ref{app:ablation}.
\section{Conclusion}

We introduce query-driven image-text joint extraction from multimodal long documents, a task that requires systems to retrieve textual attributes and corresponding images from domain-specific long documents according to user queries. To support this task, we analyze its core challenges from both user intent and the document itself, design a two-level taxonomy at the query and instance levels, and construct ITJoint, the first high-quality manually annotated benchmark for this task. We further study two representative solution settings: standalone VLMs and our task-specific multi-agent workflow, Q2IT. Experiments show that standalone VLMs struggle with this task, while Q2IT brings clear improvements but still remains far from perfect. These findings demonstrate the difficulty of query-driven image-text joint extraction and highlight its value for driving future advances in multimodal models and agent-based systems.

\section*{Limitations}

Although ITJoint covers diverse domain-specific long documents and complex image-text relations, its scale is still limited by the cost of manual annotation. Constructing the dataset requires annotators to read documents page by page, clean candidate queries, locate answer instances, and annotate image bounding boxes. This process is especially costly for set queries, where all instances satisfying the query condition must be exhaustively identified. Future work can expand the document sources, domain coverage, and dataset scale to improve the representativeness of the benchmark.

The benchmark also has several annotation-related limitations. Although ITJoint includes cross-page samples with varying spans, nearby-page cases account for a relatively large proportion, which may make the contribution of multi-strategy page search less pronounced in the overall results. In addition, some domain-specific images have ambiguous boundaries between the target object and its background. Although we use unified annotation guidelines to improve consistency, such boundary ambiguity may still affect fine-grained image localization evaluation.

Methodologically, Q2IT adopts a structured three-stage workflow that models humans' progressive focusing behavior, rather than allowing multimodal large language models to perform long-document search and image-text matching fully autonomously. This design improves controllability and traceability, but the framework still depends on the quality of intermediate results across stages. Future work can explore more flexible planning and feedback mechanisms to further improve end-to-end coordination while maintaining reliability.
\section*{Ethical Considerations}

Our dataset does not contain any personal information, and all source materials are obtained from publicly accessible channels. The dataset construction and refinement process is conducted with full respect for copyright and intellectual property rights.



\bibliography{custom}

\appendix
\clearpage

\section{Appendix}
\label{sec:appendix}

\subsection{Annotation Details}
\label{app:annotation}
\begin{figure}[t]
    \centering
    \includegraphics[width=\columnwidth]{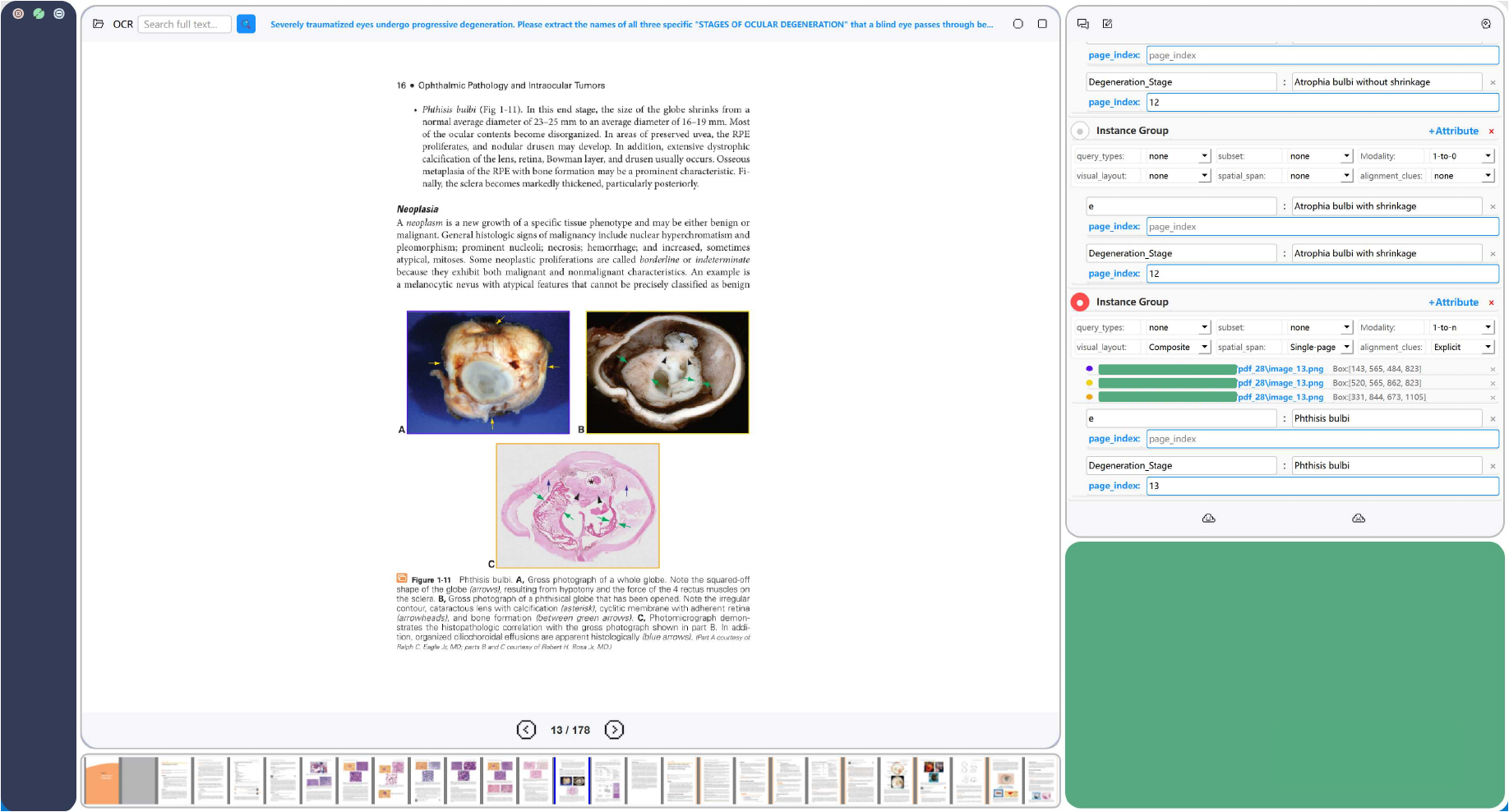}
    \caption{Interface of ITLabel.}
\label{fig:itlabel}
\end{figure}

\paragraph{ITLabel Annotation Tool.}
To improve annotation efficiency and consistency, we develop \textbf{ITLabel}, a visual annotation tool specifically designed for query-driven image-text joint extraction from long documents. As shown in Figure~\ref{fig:itlabel}, ITLabel provides a unified document-level annotation interface, rather than treating each page as an isolated image. It supports page browsing, OCR-based text extraction, keyword search, candidate-query revision, textual-attribute entry, image bounding-box annotation, and query-level and instance-level taxonomy labeling.

A key feature of ITLabel is its support for cross-page, multi-instance, and multi-image annotation. For a single query, annotators can create multiple answer instances; for a single answer instance, they can further link the textual attribute value to one or more image boxes across different pages. This design enables efficient annotation of set queries, cross-page cases, and 1-to-N image-text mappings. The OCR-based text extraction function also allows annotators to directly reuse document text when filling textual attributes, reducing manual transcription effort.

\paragraph{Document Selection Rules.}
We collect long documents from public resources to cover realistic document scenarios faced by researchers when acquiring image-text information. Document selection follows four principles: 
(1) \textbf{image richness}, where documents should contain abundant images that are indispensable for conveying core domain knowledge rather than merely decorative; 
(2) \textbf{domain diversity}, where documents should cover diverse domain-specific fields, such as medicine, agriculture, archaeology, biology, and art; 
(3) \textbf{language coverage}, where documents should cover different language settings, including English, Chinese, and bilingual documents; and 
(4) \textbf{document length}, where each document should contain a continuous page range from tens to over one hundred pages to form realistic long-range retrieval scenarios.

\paragraph{Query Generation Rules.}
We adopt a human-LLM collaborative workflow to generate queries. Specifically, we first use an LLM to generate candidate queries in batches based on the OCR text of each document. Human annotators then manually filter, revise, and supplement the candidate queries. Query generation follows seven principles: 
(1) \textbf{objectivity}, where the target attribute of each query should have a deterministic answer in the document; 
(2) \textbf{clarity}, where the query should be clearly formulated so that annotators can understand the requested information without ambiguity;
(3) \textbf{diversity}, where queries should cover diverse language styles and attribute dimensions while avoiding repetitive templates; 
(4) \textbf{non-redundancy}, where similar queries about the same entity should be avoided; 
(5) \textbf{distribution balance}, where the number of involved instances and the distribution of categories should be kept as balanced as possible; 
(6) \textbf{self-containment}, where each query should be independently understandable without relying on additional context; and 
(7) \textbf{multimodal dependency}, where each query must require joint extraction of textual attributes and corresponding images rather than purely textual question answering.
\paragraph{Answer Annotation Rules.}
For each query, annotators locate all entities satisfying the query condition in the document, record their textual attribute values, annotate image bounding boxes on the corresponding pages, and assign query-level categories and instance-level attributes. Answer annotation follows four rules: 
(1) \textbf{entity specificity}, where each answer instance must refer to a specific individual entity rather than a generic concept; 
(2) \textbf{textual faithfulness}, where textual attribute values should be directly copied from the document and should not be manually summarized or inferred without evidence; 
(3) \textbf{explicit negative annotation}, where entities with textual evidence but no corresponding image must be explicitly recorded with an empty image set; and 
(4) \textbf{independent image boxes}, where each independent visual entity should be annotated with a separate bounding box, and multiple distinct images should not be merged into one box.

For set queries, annotators are required to exhaustively identify all instances satisfying the query condition in the document. Particular attention is paid to valid instances involving cross-page relations, implicit alignment, or weak visual clues, so as to avoid missing valid answer instances.

\begin{figure}[t]
    \centering
    \includegraphics[width=\columnwidth]{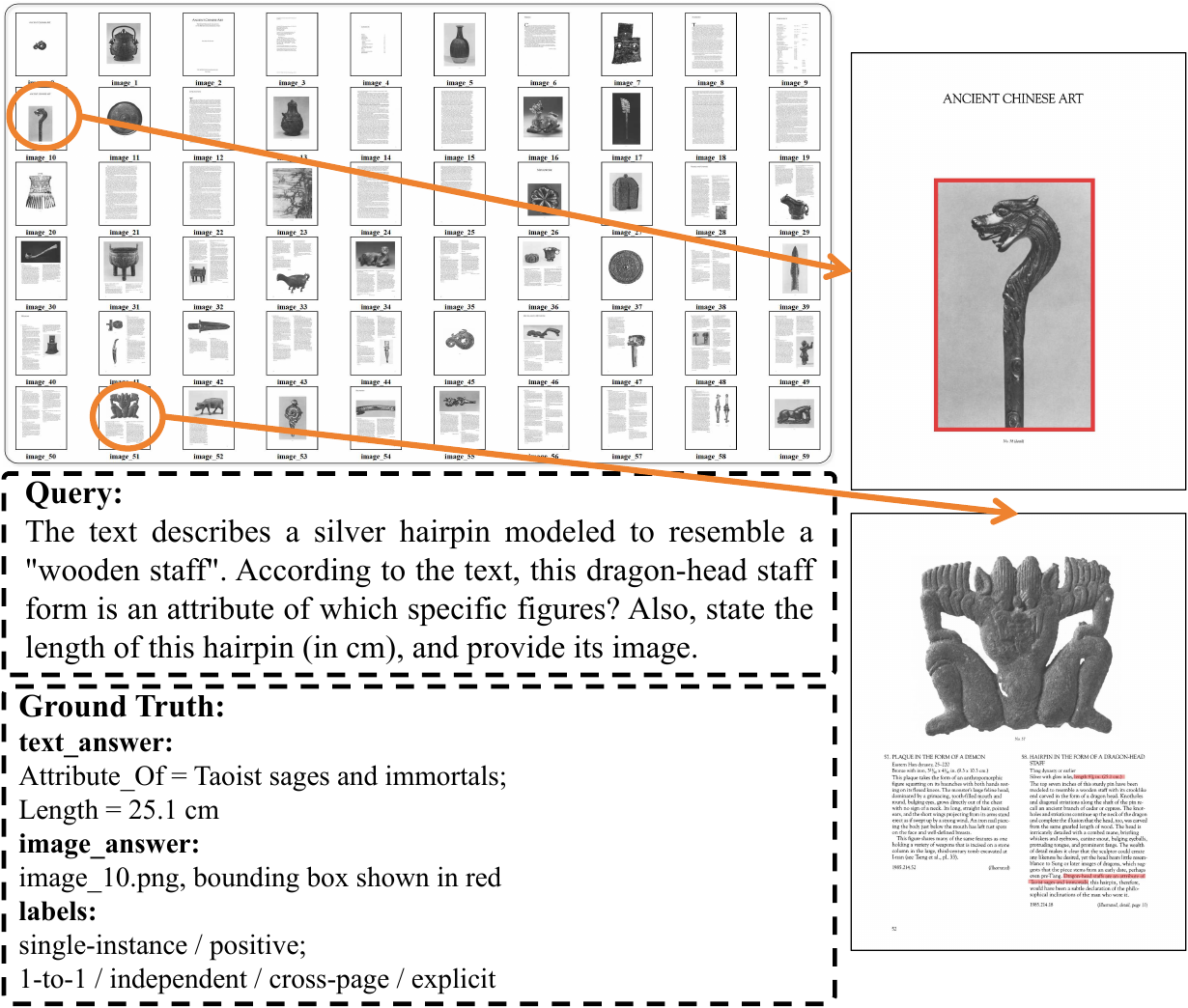}
    \caption{Ground-truth annotation example for a single-instance query in ITJoint.}
    \label{fig:ground_truth_example}
\end{figure}

\paragraph{Ground-Truth Annotation Example.}
Figure~\ref{fig:ground_truth_example} shows the ground-truth annotation format for a single-instance query in ITJoint. The query requires a textual answer and its corresponding target image. The annotation records the \texttt{text\_answer}, the \texttt{image\_answer} with the target bounding box, and the query-level and instance-level labels. This example further illustrates a cross-page case, where completing the answer instance requires information from multiple pages. For readability, we present a compact version of the annotation and omit implementation-specific fields such as local file paths.

\subsection{Instance-Level Taxonomy Examples}
\label{app:taxonomy_examples}

\begin{figure}[t]
    \centering
    \includegraphics[width=\columnwidth]{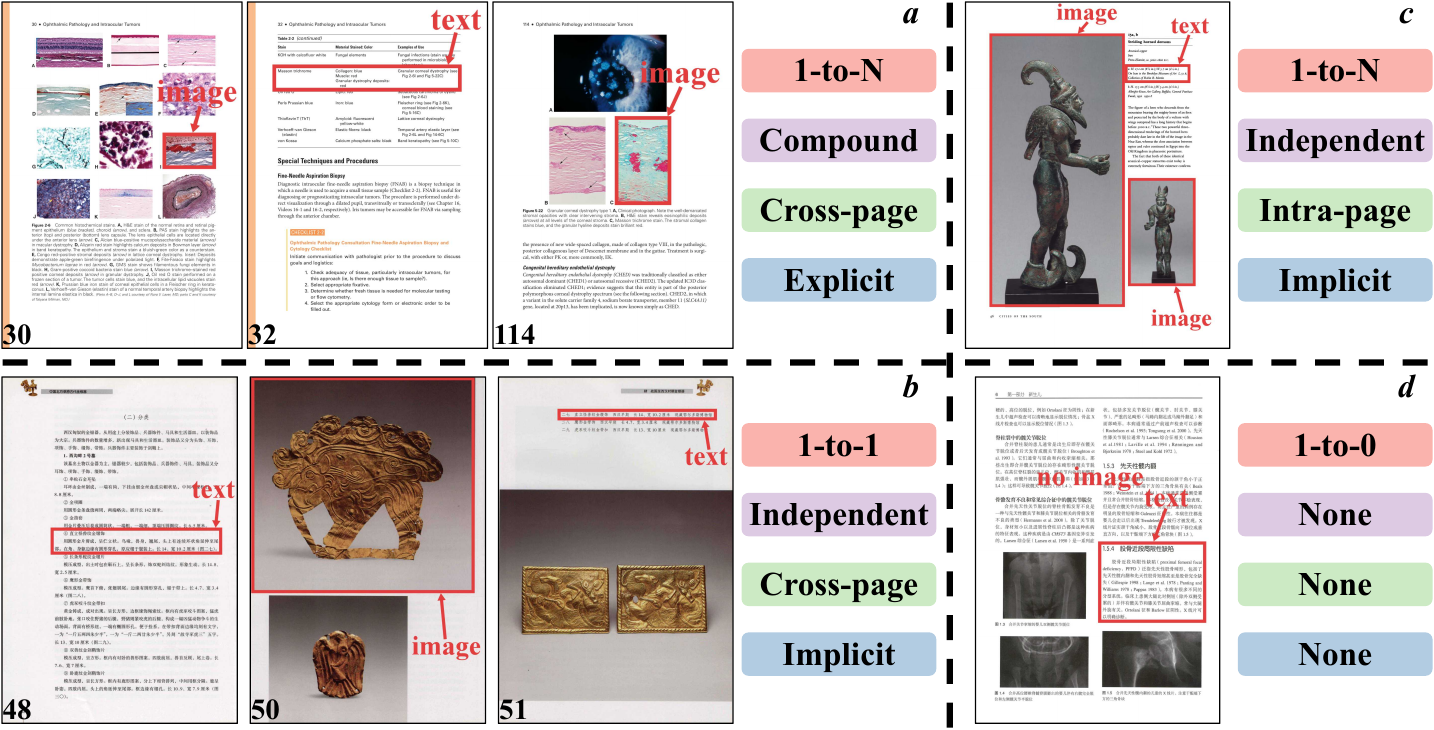}
    \caption{Instance-level taxonomy examples in ITJoint.}
    \label{fig:taxonomy_examples}
\end{figure}
Figure~\ref{fig:taxonomy_examples} provides visual examples of the instance-level taxonomy in ITJoint. Each panel corresponds to one answer instance and shows its labels along modality cardinality, visual layout, spatial span, and alignment clues. Red boxes indicate relevant textual evidence or target image regions.

Panel (a) shows an answer instance associated with multiple target images, and is therefore labeled as 1-to-N. The textual evidence and all target images together involve multiple pages, so the instance is also labeled as cross-page. Since the target images are nested in compound visual layouts and the document provides explicit label references linking the text to the images, it is further labeled as compound and explicit.

Panel (b) shows a 1-to-1 instance, where one textual attribute value corresponds to a single target image. Although the target image itself is independent, the textual evidence and the image together involve multiple pages, requiring cross-page association. The correspondence is implicit because the image-text relation cannot be established solely through an explicit marker and must be inferred from the surrounding context.

Panel (c) illustrates an intra-page instance, where the textual evidence and all target images together involve only a single page. The target images are standalone visual regions rather than parts of a compound figure, so the instance is labeled as independent. Since multiple target image regions are associated with the same textual attribute value, it is labeled as 1-to-N. Its alignment is implicit, as the association relies on page layout and local context rather than a direct textual reference.

Panel (d) shows a 1-to-0 instance. The document contains textual evidence for the target instance, but no corresponding image is present. Therefore, the system is expected to return the textual attribute value with an empty image set, i.e., $v_j=\emptyset$, rather than hallucinating an image region. Since no image is associated with the instance, the visual layout, spatial span, and alignment-clue labels are not applicable.

\subsection{Document Examples}
\label{app:document_examples}

As described in the document collection process, ITJoint collects long documents from multiple domain-specific fields, including medicine, agriculture, archaeology, biology, and art. These documents cover both English and Chinese, contain diverse image-text layouts, and include rich non-decorative core images. We show examples of these documents below.
\begin{figure*}[t]
    \centering
    \includegraphics[width=\textwidth]{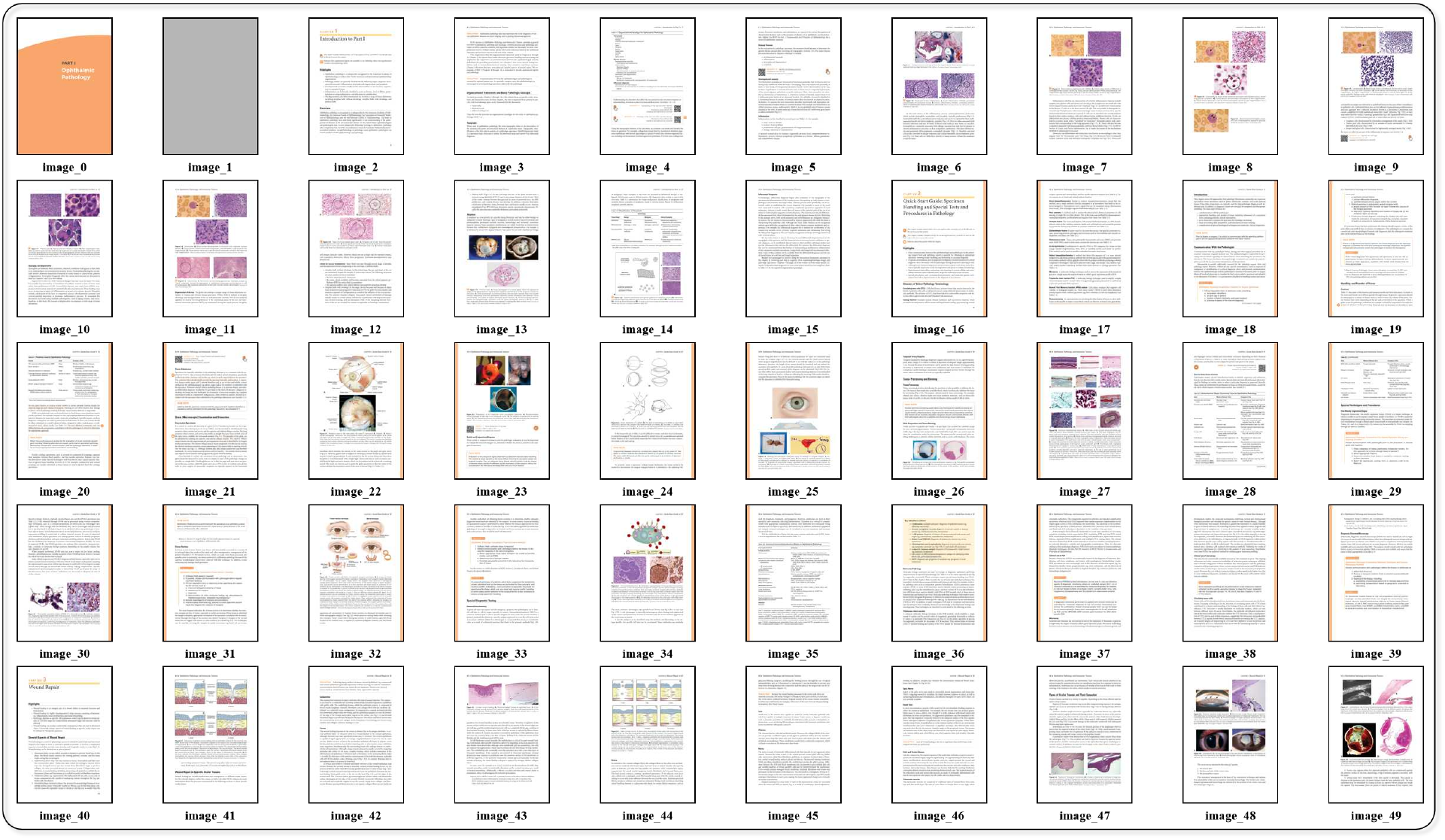}
    \caption{Medical document examples from ITJoint (English).}
    \label{fig:28}
\end{figure*}

\begin{figure*}[t]
    \centering
    \includegraphics[width=\textwidth]{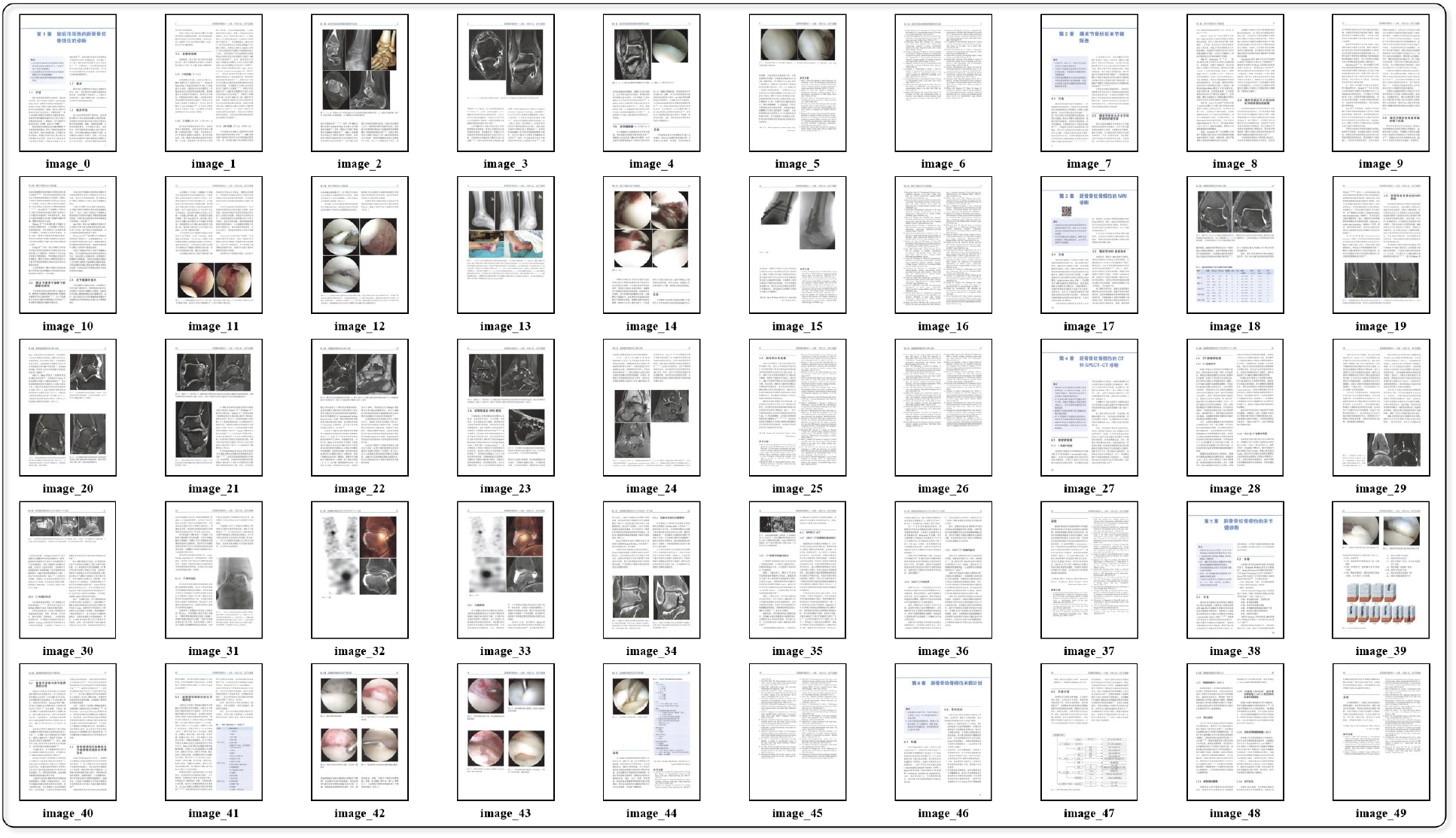}
    \caption{Medical document examples from ITJoint (Chinese).}
    \label{fig:5}
\end{figure*}

\begin{figure*}[t]
    \centering
    \includegraphics[width=\textwidth]{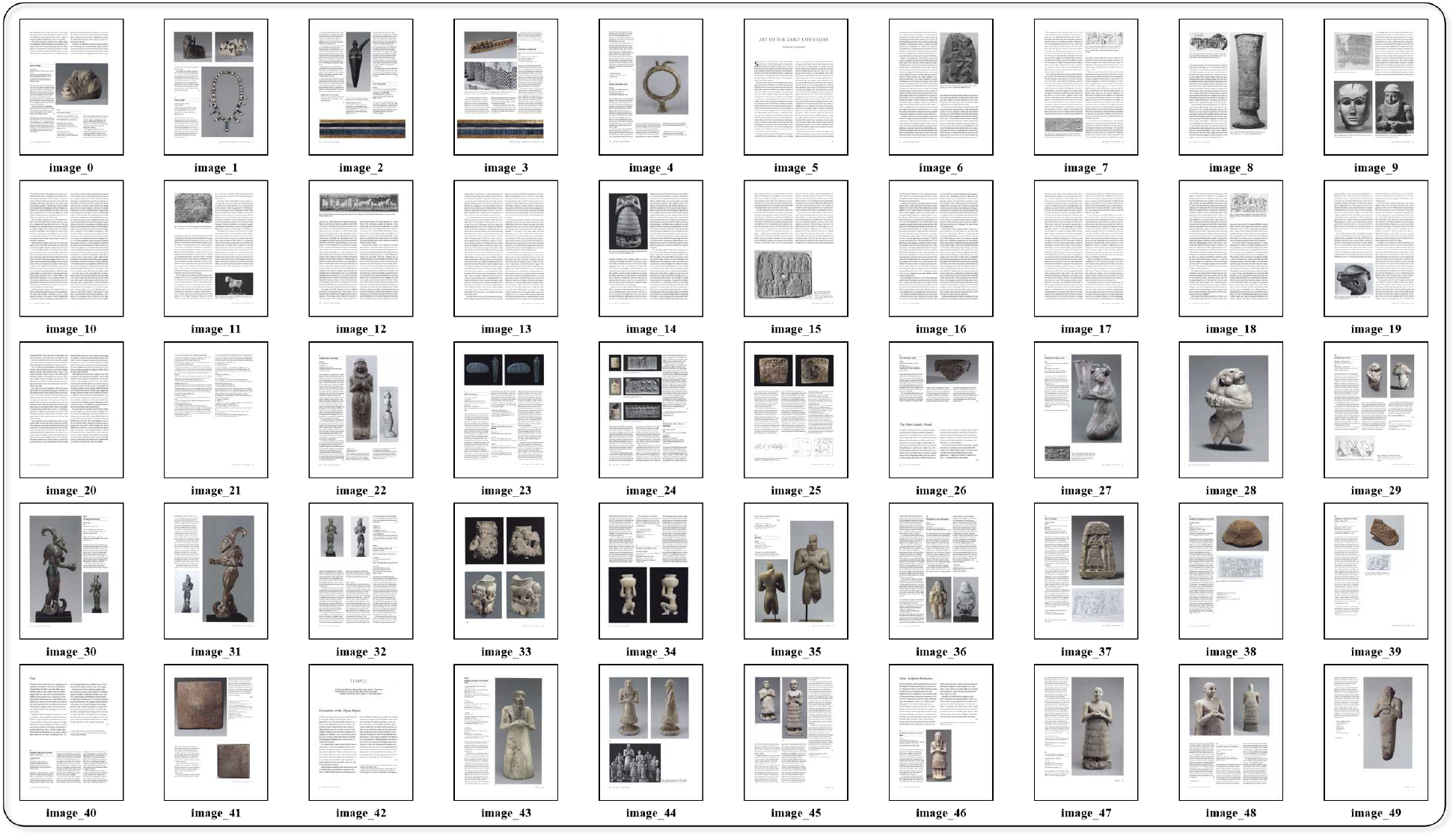}
    \caption{Archaeological document examples from ITJoint (English).}
    \label{fig:40}
\end{figure*}

\begin{figure*}[t]
    \centering
    \includegraphics[width=\textwidth]{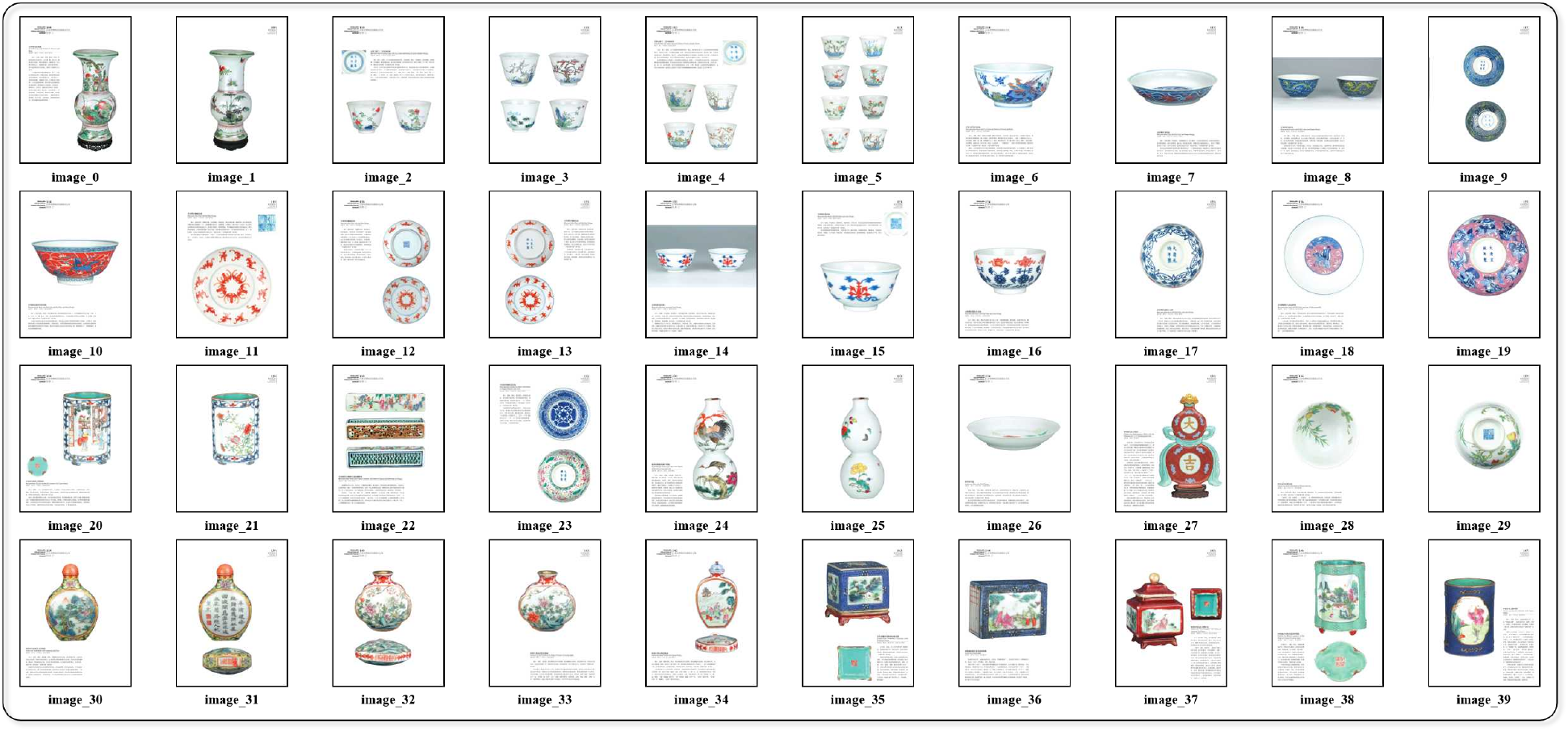}
    \caption{Archaeological document examples from ITJoint (Chinese).}
    \label{fig:0}
\end{figure*}

\begin{figure*}[t]
    \centering
    \includegraphics[width=\textwidth]{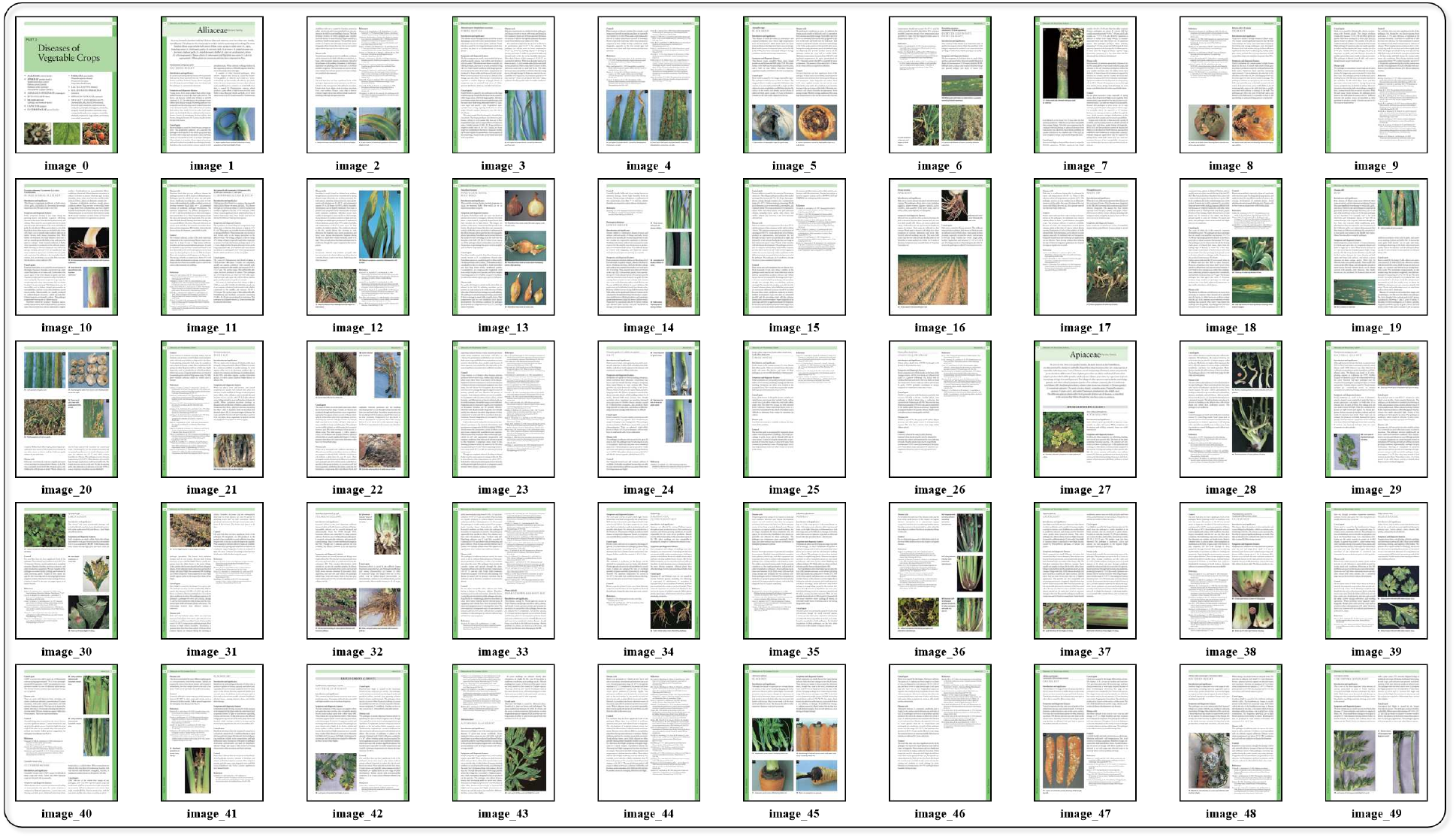}
    \caption{Agricultural document examples from ITJoint (English).}
    \label{fig:16}
\end{figure*}

\begin{figure*}[t]
    \centering
    \includegraphics[width=\textwidth]{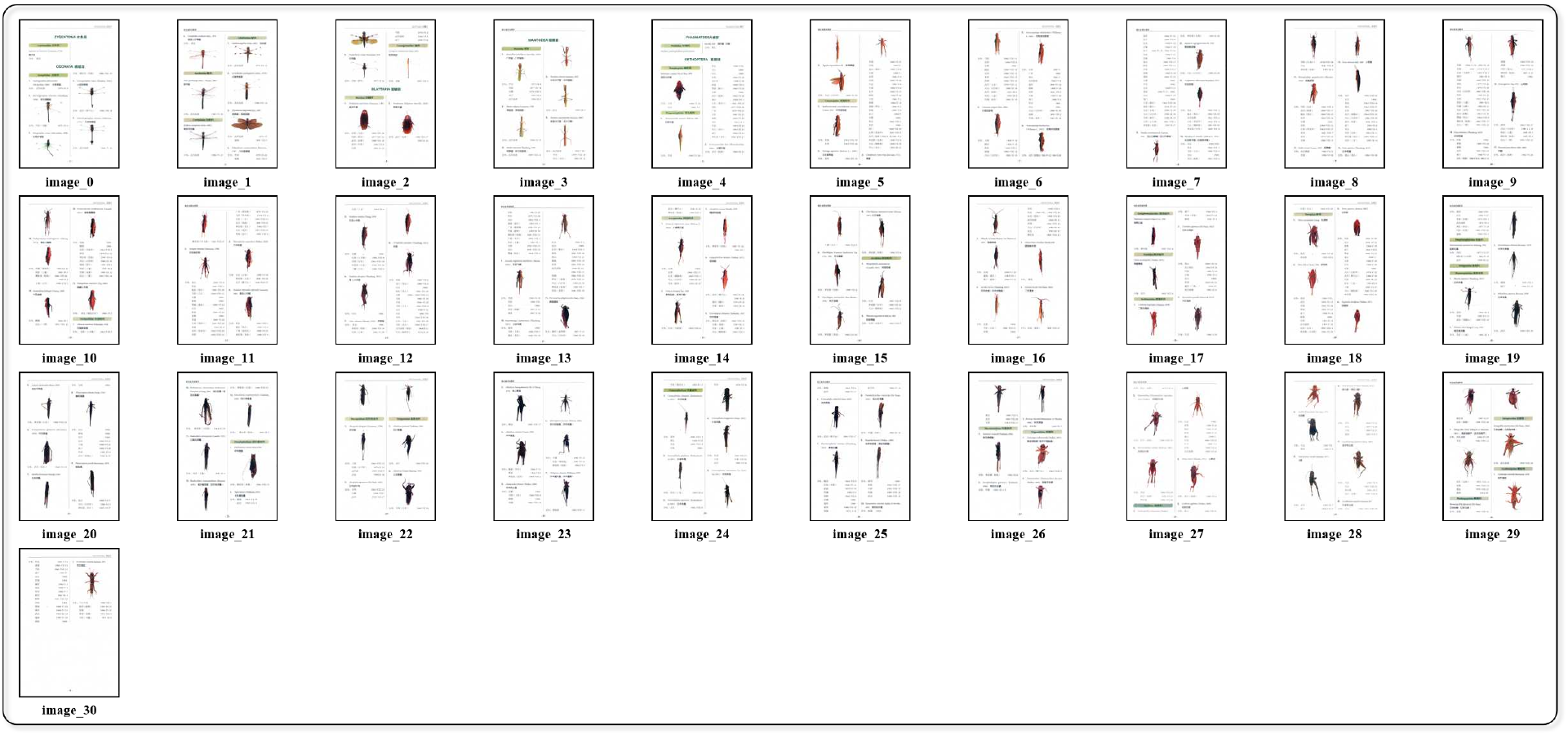}
    \caption{Biology document examples from ITJoint (Chinese-English).}
    \label{fig:12}
\end{figure*}

\begin{figure*}[t]
    \centering
    \includegraphics[width=\textwidth]{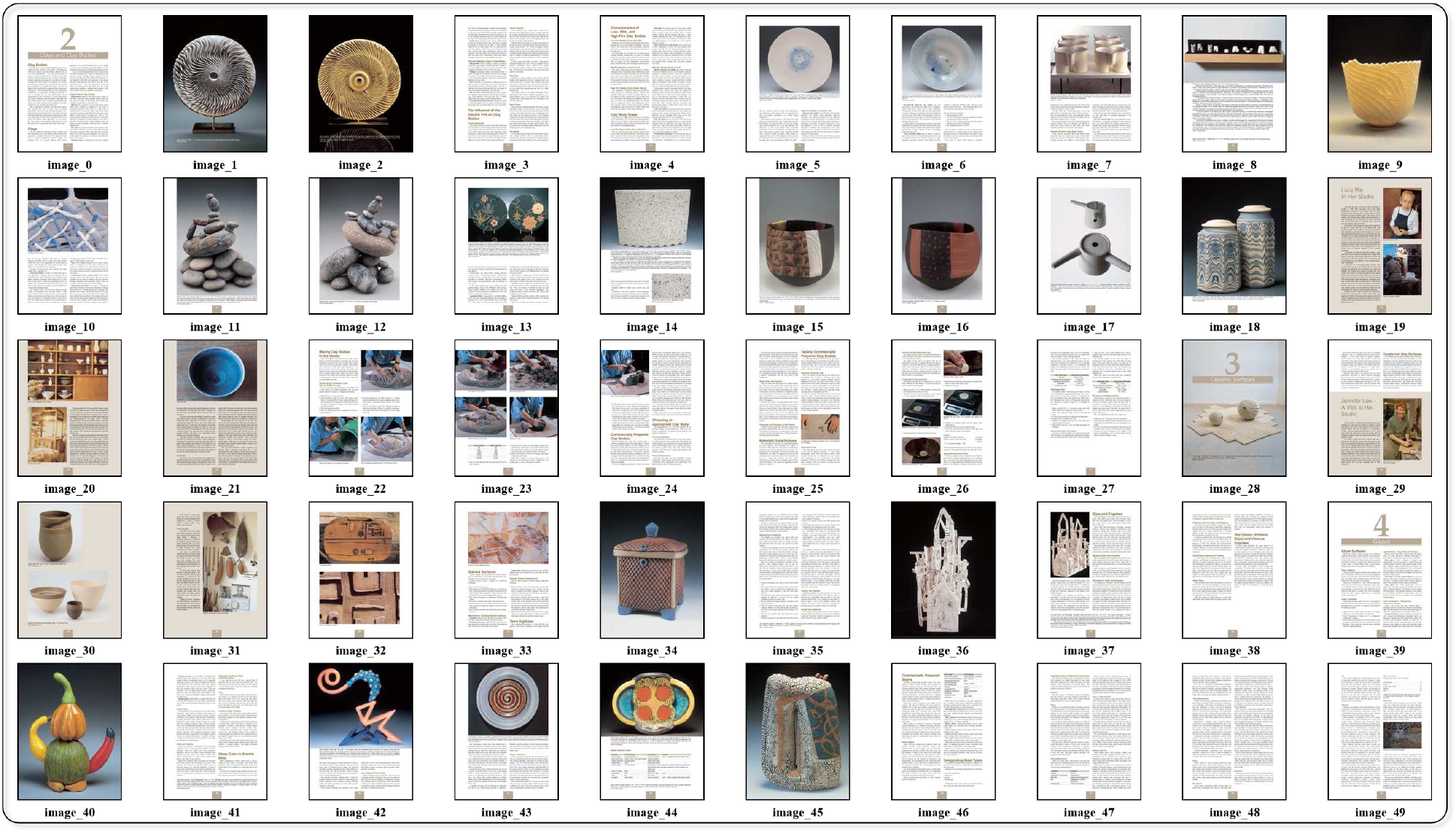}
    \caption{Art document examples from ITJoint (English).}
    \label{fig:15}
\end{figure*}

\begin{figure*}[t]
    \centering
    \includegraphics[width=\textwidth]{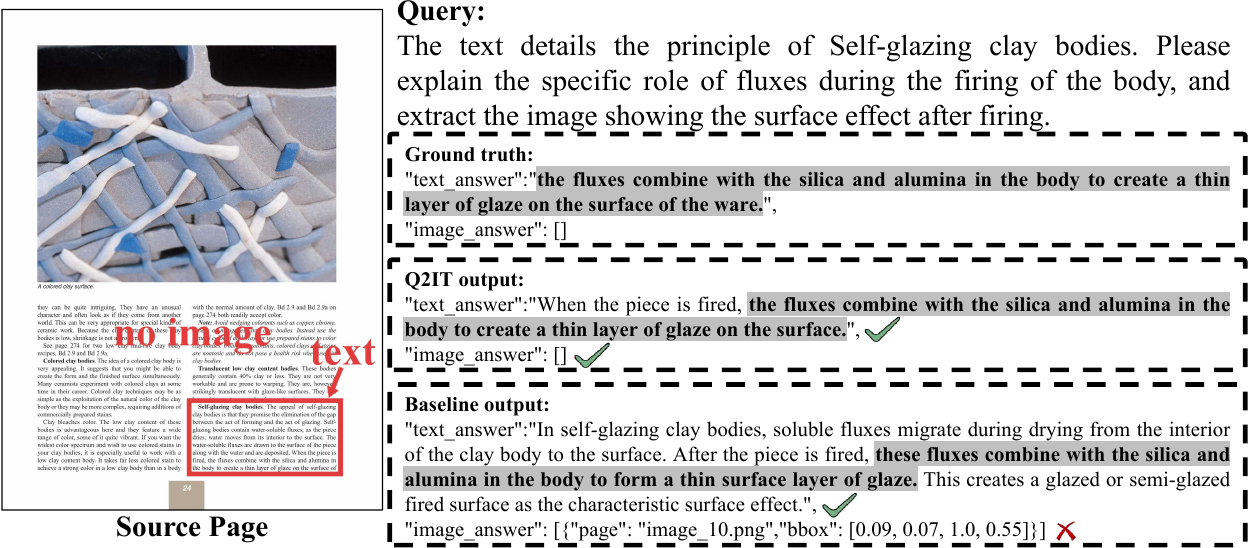}
    \caption{Qualitative example of image-text alignment hallucination.}
    \label{fig:hallucination_example}
\end{figure*}

\subsection{Qualitative Example of Image-Text Alignment Hallucination}
\label{app:hallucination_example}
Figure~\ref{fig:hallucination_example} shows a qualitative example of image-text alignment hallucination on a 1-to-0 instance using GPT as the backbone. The document contains textual evidence for the queried entity, but no corresponding target image exists. Under direct inference, GPT correctly identifies the textual answer but still assigns an image region, causing the instance to be correct under text accuracy but incorrect under joint accuracy. In contrast, GPT equipped with Q2IT returns the same textual answer with an empty image set, which is consistent with the ground truth.

\subsection{Additional Results under the Soft Joint Setting}
\label{app:soft_results}
In the main experiments, we report results under the Strict Joint setting, which requires the predicted image set to exactly match the gold image set. This setting is appropriate for evaluating complete image-text joint extraction, but it may be overly conservative in scenarios where users do not necessarily need all corresponding images to complete the downstream task. In such cases, returning a subset of correct images can still be useful, as long as the system does not introduce incorrect or hallucinated image regions.

Therefore, we additionally report results under the Soft Joint setting in Table~\ref{tab:soft_results}. Soft Joint relaxes the image-box cardinality constraint by allowing the prediction to contain fewer boxes than the gold instance. For image-bearing instances, the prediction must contain at least one correctly matched box, and every predicted box must match a gold box of the same instance. For 1-to-0 instances, the prediction is still required to return an empty image set. Thus, Soft Joint rewards partially complete but correct image retrieval, while still penalizing irrelevant or hallucinated images. Compared with the Strict Joint results in the main experiments, Soft Joint generally yields higher scores, especially for 1-to-N instances. Nevertheless, the overall trend remains consistent: Q2IT improves all backbone models across query-level and instance-level categories, indicating that the effectiveness of our framework is robust across evaluation settings.

\subsection{Implementation Details}
\label{app:implementation_details}

We use Qwen3-Embedding-8B to encode page-indexed text chunks. For hybrid retrieval, FAISS and BM25 each return the top 30 candidates, which are fused and then reranked by BAAI/bge-reranker-v2-m3. The top 10 reranked chunks are used for evidence extraction. For section-restricted queries, after detecting a section title, Q2IT expands a 30-page window based on page metadata; if page indices cannot be reliably parsed, it falls back to the next 50 chunks.

\subsection{Ablation Study}
\label{app:ablation}

Unless otherwise specified, all ablation experiments are conducted with GPT-5.4 as the backbone model under the Q2IT framework. The results are reported in Table~\ref{tab:ablation_results}.

\paragraph{Page Agent ablation.}
We ablate two key designs in the Page Agent: the Page Checker and the multi-strategy page search. Removing the Page Checker causes the most significant drop in 1-to-0 accuracy, from 0.802 to 0.531, a decrease of 27.1 percentage points, indicating that this module is crucial for suppressing image-text alignment hallucinations. Accuracy on cross-page and implicit-alignment instances also drops by more than 10 percentage points, while the declines for intra-page and explicit-alignment instances are relatively mild. This comparison verifies our observation that when a target image is far from its descriptive text, long documents usually provide localizable clues, such as figure labels, printed page numbers, or relative page references. When such clues are absent, the target image is often located on the evidence page or nearby pages. Explicit references usually provide captions or labels and can be localized through search, making them less dependent on the Page Checker; implicit cross-page images lack such anchors and therefore rely more on visual verification.

We further remove the multi-strategy page search after removing the Page Checker, using only the source pages output by the Evidence Agent as candidate pages. Page-level F1 drops from 0.799 to 0.740, and cross-page accuracy further decreases from 0.349 to 0.234. This confirms that target images in long documents are not always located on the same pages as their descriptive text, and source pages cannot replace dedicated page search. Accuracy on 1-to-N instances also drops further, from 0.355 to 0.307, because multiple images corresponding to the same entity are more likely to be distributed across different pages; without page search, some target-image pages cannot be located.

\paragraph{External tool ablation.}
We examine the necessity of using an external open-set detector for precise image-region extraction. Removing Grounding DINO leads to large drops across image-related instance categories: accuracy on 1-to-1 instances decreases from 0.534 to 0.170, accuracy on 1-to-N instances decreases from 0.452 to 0.078, and accuracy on compound instances decreases from 0.387 to 0.073. This shows that current multimodal models still lack sufficient direct coordinate-generation ability for precise region extraction, making a dedicated open-set object detector necessary for this task.

\begin{table*}[t]
\centering
\caption{Additional results of direct inference and Q2IT on ITJoint under the Soft Joint setting.}
\label{tab:soft_results}
\scriptsize
\setlength{\tabcolsep}{3.6pt}
\renewcommand{\arraystretch}{1.12}
\resizebox{\textwidth}{!}{
\begin{tabular}{lccccccccccccc}
\toprule
\textbf{Model}
& \textbf{Single} & \textbf{Single} & \textbf{Set} & \textbf{Set}
& \textbf{1-to-0} & \textbf{1-to-1} & \textbf{1-to-N}
& \textbf{Indep.} & \textbf{Comp.}
& \textbf{Intra} & \textbf{Cross}
& \textbf{Exp.} & \textbf{Imp.} \\
& \textbf{Pos.} & \textbf{Neg.} & \textbf{Pos.} & \textbf{Neg.}
& & & & & & & & & \\
\midrule
\multicolumn{14}{c}{\textbf{Closed-source Models}} \\
\midrule
Gemini-3.1-Pro-Preview
& 0.466 & 0.714 & 0.410 & 0.545
& 0.744 & 0.482 & 0.333
& 0.509 & 0.183
& 0.505 & 0.364
& 0.458 & 0.429 \\

Gemini-3.1-Pro-Preview (Q2IT)
& \textbf{0.727}\gain{+0.261}
& \textbf{0.943}\gain{+0.229}
& 0.583\gain{+0.173}
& \textbf{0.722}\gain{+0.177}
& \textbf{0.876}\gain{+0.132}
& \textbf{0.563}\gain{+0.081}
& 0.596\gain{+0.263}
& \textbf{0.592}\gain{+0.083}
& \textbf{0.475}\gain{+0.292}
& 0.550\gain{+0.046}
& \textbf{0.600}\gain{+0.236}
& 0.607\gain{+0.149}
& \textbf{0.498}\gain{+0.069} \\

GPT-5.4
& 0.099 & 0.393 & 0.107 & 0.310
& 0.541 & 0.113 & 0.042
& 0.118 & 0.007
& 0.091 & 0.105
& 0.080 & 0.131 \\

GPT-5.4 (Q2IT)
& 0.649\gain{+0.550}
& 0.805\gain{+0.412}
& \textbf{0.624}\gain{+0.517}
& 0.701\gain{+0.391}
& 0.802\gain{+0.262}
& 0.536\gain{+0.423}
& \textbf{0.621}\gain{+0.578}
& 0.579\gain{+0.461}
& 0.453\gain{+0.445}
& \textbf{0.555}\gain{+0.464}
& 0.556\gain{+0.451}
& \textbf{0.623}\gain{+0.543}
& 0.420\gain{+0.290} \\

\midrule
\multicolumn{14}{c}{\textbf{Open-source Models}} \\
\midrule
Qwen3-VL-235B-A22B-Instruct
& 0.269 & 0.309 & 0.256 & 0.305
& 0.387 & 0.230 & 0.227
& 0.258 & 0.101
& 0.265 & 0.171
& 0.258 & 0.176 \\

Qwen3-VL-235B-A22B-Instruct (Q2IT)
& 0.509\gain{+0.240}
& 0.623\gain{+0.314}
& 0.476\gain{+0.220}
& 0.425\gain{+0.119}
& 0.528\gain{+0.142}
& 0.382\gain{+0.152}
& \textbf{0.473}\gain{+0.246}
& \textbf{0.424}\gain{+0.166}
& 0.309\gain{+0.208}
& 0.406\gain{+0.141}
& \textbf{0.396}\gain{+0.225}
& 0.444\gain{+0.187}
& \textbf{0.318}\gain{+0.143} \\

GLM-4.5V
& 0.074 & 0.304 & 0.056 & 0.158
& 0.273 & 0.034 & 0.102
& 0.057 & 0.015
& 0.048 & 0.051
& 0.045 & 0.057 \\

GLM-4.5V (Q2IT)
& \textbf{0.535}\gain{+0.460}
& \textbf{0.752}\gain{+0.448}
& \textbf{0.490}\gain{+0.434}
& \textbf{0.580}\gain{+0.422}
& \textbf{0.644}\gain{+0.371}
& \textbf{0.388}\gain{+0.355}
& 0.452\gain{+0.349}
& 0.416\gain{+0.359}
& \textbf{0.343}\gain{+0.329}
& \textbf{0.418}\gain{+0.370}
& 0.380\gain{+0.329}
& \textbf{0.459}\gain{+0.414}
& 0.290\gain{+0.233} \\
\bottomrule
\end{tabular}
}
\end{table*}

\begin{table*}[t]
\centering
\caption{Ablation results of Q2IT on ITJoint with GPT-5.4 as the backbone model.}
\label{tab:ablation_results}
\scriptsize
\setlength{\tabcolsep}{3.6pt}
\renewcommand{\arraystretch}{1.12}
\resizebox{\textwidth}{!}{
\begin{tabular}{lcccccccccc}
\toprule
\textbf{Setting}
& \textbf{1-to-0}
& \textbf{1-to-1}
& \textbf{1-to-N}
& \textbf{Indep.}
& \textbf{Comp.}
& \textbf{Intra}
& \textbf{Cross}
& \textbf{Exp.}
& \textbf{Imp.}
& \textbf{Page} \\
\midrule
\multicolumn{11}{c}{\textbf{Page Agent Ablation}} \\
\midrule
Full Q2IT
& \textbf{0.802} & \textbf{0.534} & \textbf{0.452} & \textbf{0.545} & \textbf{0.387} & \textbf{0.530} & \textbf{0.495} & \textbf{0.574} & \textbf{0.400} & \textbf{0.872} \\
w/o Page Checker
& 0.531\drop{0.271} & 0.457\drop{0.078} & 0.355\drop{0.096} & 0.453\drop{0.092} & 0.350\drop{0.037} & 0.491\drop{0.039} & 0.349\drop{0.146} & 0.510\drop{0.064} & 0.282\drop{0.118} & 0.799\drop{0.072} \\
w/o Page Search
& 0.548\drop{0.254} & 0.436\drop{0.099} & 0.307\drop{0.145} & 0.431\drop{0.114} & 0.299\drop{0.088} & 0.523\drop{0.007} & 0.234\drop{0.261} & 0.469\drop{0.105} & 0.282\drop{0.118} & 0.740\drop{0.132} \\
\midrule
\multicolumn{11}{c}{\textbf{External Tool Ablation}} \\
\midrule
Full Q2IT
& \textbf{0.802} & \textbf{0.534} & \textbf{0.452} & \textbf{0.545} & \textbf{0.387} & \textbf{0.530} & \textbf{0.495} & \textbf{0.574} & \textbf{0.400} & 0.872 \\
w/o Grounding DINO
& 0.774\drop{0.028} & 0.170\drop{0.365} & 0.078\drop{0.374} & 0.166\drop{0.379} & 0.073\drop{0.314} & 0.151\drop{0.379} & 0.146\drop{0.349} & 0.121\drop{0.453} & 0.204\drop{0.196} & \textbf{0.888}\gain{+0.017} \\
\bottomrule
\end{tabular}
}
\end{table*}
\end{document}